\documentclass{article}

\usepackage{microtype}
\usepackage{graphicx}
\usepackage{subfigure}
\usepackage{booktabs} 
\PassOptionsToPackage{hyphens}{url}\usepackage{hyperref}

\makeatletter
\g@addto@macro{\UrlBreaks}{\UrlOrds}
\makeatother

\usepackage[accepted]{icml2018}

\usepackage[utf8]{inputenc}
\usepackage[T1]{fontenc}
\usepackage{url}
\usepackage{booktabs}
\usepackage{amsfonts}
\usepackage{nicefrac}
\usepackage{microtype}      

\usepackage{setspace}

\usepackage{endnotes}
\usepackage{subfigure}                  \usepackage{amsmath}                     
\usepackage{amssymb}                     
\usepackage{amsfonts}                            
\usepackage[american]{babel}
\usepackage{graphicx}

\usepackage[dvipsnames]{xcolor}

\usepackage{pgf,tikz}
\usepackage{pgfplots}
\usepackage{pgfplotstable}
\usetikzlibrary{calc,shadows}
\usetikzlibrary{pgfplots.groupplots}
\usepackage{subfigure}
\usepackage{mathtools} \usepackage{mathtools}
\usepackage{array,multirow,graphicx}

\pgfplotsset{select coords between index/.style 2 args={
    x filter/.code={
        \ifnum\coordindex<#1\fi
        \ifnum\coordindex>#2\fi
    }
}}

\pgfplotstableread{mul.txt}{\toponemul}
\pgfplotstablecreatecol[copy column from table={top1.txt}{0}] {topone} {\toponemul}
\pgfplotstablecreatecol[copy column from table={color.txt}{0}] {class} {\toponemul}

\pgfplotstableread{add.txt}{\toponeadd}
\pgfplotstablecreatecol[copy column from table={top1.txt}{0}] {topone} {\toponeadd}
\pgfplotstablecreatecol[copy column from table={color.txt}{0}] {class} {\toponeadd}

\pgfplotstableread{space.txt}{\toponespace}
\pgfplotstablecreatecol[copy column from table={top1.txt}{0}] {topone} {\toponespace}
\pgfplotstablecreatecol[copy column from table={color.txt}{0}] {class} {\toponespace}

\pgfplotstableread{mul.txt}{\topfivemul}
\pgfplotstablecreatecol[copy column from table={top5.txt}{0}] {topone} {\topfivemul}
\pgfplotstablecreatecol[copy column from table={color.txt}{0}] {class} {\topfivemul}

\pgfplotstableread{add.txt}{\topfiveadd}
\pgfplotstablecreatecol[copy column from table={top5.txt}{0}] {topone} {\topfiveadd}
\pgfplotstablecreatecol[copy column from table={color.txt}{0}] {class} {\topfiveadd}

\pgfplotstableread{space.txt}{\topfivespace}
\pgfplotstablecreatecol[copy column from table={top5.txt}{0}] {topone} {\topfivespace}
\pgfplotstablecreatecol[copy column from table={color.txt}{0}] {class} {\topfivespace}

\pgfplotstableread{mul_dist.txt}{\toponemuldist}
\pgfplotstablecreatecol[copy column from table={top1_dist.txt}{0}] {topone} {\toponemuldist}
\pgfplotstablecreatecol[copy column from table={color_dist.txt}{0}] {class} {\toponemuldist}

\pgfplotstableread{add_dist.txt}{\toponeadddist}
\pgfplotstablecreatecol[copy column from table={top1_dist.txt}{0}] {topone} {\toponeadddist}
\pgfplotstablecreatecol[copy column from table={color_dist.txt}{0}] {class} {\toponeadddist}

\pgfplotstableread{space_dist.txt}{\toponespacedist}
\pgfplotstablecreatecol[copy column from table={top1_dist.txt}{0}] {topone} {\toponespacedist}
\pgfplotstablecreatecol[copy column from table={color_dist.txt}{0}] {class} {\toponespacedist}

\pgfplotstableread{mul_dist.txt}{\topfivemuldist}
\pgfplotstablecreatecol[copy column from table={top5_dist.txt}{0}] {topone} {\topfivemuldist}
\pgfplotstablecreatecol[copy column from table={color_dist.txt}{0}] {class} {\topfivemuldist}

\pgfplotstableread{add_dist.txt}{\topfiveadddist}
\pgfplotstablecreatecol[copy column from table={top5_dist.txt}{0}] {topone} {\topfiveadddist}
\pgfplotstablecreatecol[copy column from table={color_dist.txt}{0}] {class} {\topfiveadddist}

\pgfplotstableread{space_dist.txt}{\topfivespacedist}
\pgfplotstablecreatecol[copy column from table={top5_dist.txt}{0}] {topone} {\topfivespacedist}
\pgfplotstablecreatecol[copy column from table={color_dist.txt}{0}] {class} {\topfivespacedist}

\pgfplotstableread{mul_lm.txt}{\ppwmul}
\pgfplotstablecreatecol[copy column from table={ppw_lm.txt}{0}] {ppw} {\ppwmul}
\pgfplotstablecreatecol[copy column from table={ppw_lm.txt}{1}] {err} {\ppwmul}
\pgfplotstablecreatecol[copy column from table={color_lm.txt}{0}] {class} {\ppwmul}

\pgfplotstableread{add_lm.txt}{\ppwadd}
\pgfplotstablecreatecol[copy column from table={ppw_lm.txt}{0}] {ppw} {\ppwadd}
\pgfplotstablecreatecol[copy column from table={ppw_lm.txt}{1}] {err} {\ppwadd}
\pgfplotstablecreatecol[copy column from table={color_lm.txt}{0}] {class} {\ppwadd}

\pgfplotstableread{space_lm.txt}{\ppwspace}
\pgfplotstablecreatecol[copy column from table={ppw_lm.txt}{0}] {ppw} {\ppwspace}
\pgfplotstablecreatecol[copy column from table={ppw_lm.txt}{1}] {err} {\ppwspace}
\pgfplotstablecreatecol[copy column from table={color_lm.txt}{0}] {class} {\ppwspace}

\pgfplotstableread{mul_lm.txt}{\ppwmuldist}
\pgfplotstablecreatecol[copy column from table={ppw_lm_dist.txt}{0}] {ppw} {\ppwmuldist}
\pgfplotstablecreatecol[copy column from table={ppw_lm_dist.txt}{1}] {err} {\ppwmuldist}
\pgfplotstablecreatecol[copy column from table={color_lm.txt}{0}] {class} {\ppwmuldist}

\pgfplotstableread{add_lm.txt}{\ppwadddist}
\pgfplotstablecreatecol[copy column from table={ppw_lm_dist.txt}{0}] {ppw} {\ppwadddist}
\pgfplotstablecreatecol[copy column from table={ppw_lm_dist.txt}{1}] {err} {\ppwadddist}
\pgfplotstablecreatecol[copy column from table={color_lm.txt}{0}] {class} {\ppwadddist}

\pgfplotstableread{space_lm.txt}{\ppwspacedist}
\pgfplotstablecreatecol[copy column from table={ppw_lm_dist.txt}{0}] {ppw} {\ppwspacedist}
\pgfplotstablecreatecol[copy column from table={ppw_lm_dist.txt}{1}] {err} {\ppwspacedist}
\pgfplotstablecreatecol[copy column from table={color_lm.txt}{0}] {class} {\ppwspacedist}

\newcommand{\R}{\mathbb R}
\newcommand{\K}{\mathbb K}
\renewcommand{\vec}{\mathrm{vec}}
\newcommand{\cin}{c_\mathrm{in}}
\newcommand{\cout}{c_\mathrm{out}}
\newcommand{\rank}{\mathrm{rank}}
\newcommand{\nout}{n_\mathrm{out}}

\allowdisplaybreaks

\icmltitlerunning{StrassenNets: Deep Learning with a Multiplication Budget}

\usepackage[section]{placeins}
\usepackage{natbib}

\begin{document}

\twocolumn[
\icmltitle{StrassenNets: Deep Learning with a Multiplication Budget}

\icmlsetsymbol{equal}{*}

\begin{icmlauthorlist}
\icmlauthor{Michael Tschannen}{eth}
\icmlauthor{Aran Khanna}{amzn}
\icmlauthor{Anima Anandkumar}{amzn,calt}
\end{icmlauthorlist}

\icmlaffiliation{eth}{ETH Z{\"u}rich, Z{\"u}rich, Switzerland (most of this work was done while MT was at Amazon AI)}
\icmlaffiliation{amzn}{Amazon AI, Palo Alto, CA, USA}
\icmlaffiliation{calt}{Caltech, Pasadena, CA, USA}

\icmlcorrespondingauthor{Michael Tschannen}{michaelt@nari.ee.ethz.ch}

\icmlkeywords{Machine Learning, ICML}

\vskip 0.3in
]

\printAffiliationsAndNotice{}  
\begin{abstract}
A large fraction of the arithmetic operations required to evaluate deep neural networks (DNNs) consists of matrix multiplications,  in both convolution and fully connected layers. We perform end-to-end learning of low-cost approximations of matrix multiplications in DNN layers by casting  matrix multiplications  as $2$-layer sum-product networks (SPNs) (arithmetic circuits) and learning their (ternary) edge weights from data. 
The SPNs disentangle multiplication and addition operations and enable us to impose a budget on the number of multiplication operations. 
Combining our method with knowledge distillation and applying it to image classification DNNs (trained on ImageNet) and language modeling DNNs (using LSTMs), we obtain a first-of-a-kind reduction in number of multiplications (over 99.5\%) while maintaining the predictive performance of the full-precision models.
Finally, we demonstrate that the proposed framework is able to rediscover Strassen's matrix multiplication algorithm, learning to multiply $2 \times 2$ matrices using only $7$ multiplications instead of~$8$.
\end{abstract}

\section{Introduction}

The outstanding predictive performance of deep neural networks (DNNs)
often comes at the cost of large model size,
and corresponding computational inefficiency. This can make the deployment of DNNs on mobile and embedded hardware challenging.
For example, a full-precision ResNet-152 \cite{he2016deep} contains 60.2 million parameters and 
one forward pass requires 11.3 billion floating point operations. 
A variety of methods to address this issue were proposed recently, including optimizing the network architecture, factorizing the weight tensors, pruning the weights, and reducing the numerical precision of weights and activations (see Section \ref{sec:relwork} for a detailed overview). 

These prior works mainly focused on decreasing the number of multiply-accumulate operations used by DNNs. 
In contrast, in this paper, the objective that guides our algorithm design is a \emph{reduction of the number of multiplications}. 
This algorithm design principle has led to many fast algorithms in linear algebra, most notably Strassen's matrix multiplication algorithm \cite{strassen1969gaussian}. Strassen's algorithm uses $7$ instead $8$ multiplications to compute the product of two $2\times2$ matrices (and requires $O(n^{2.807})$ operations for multiplying $n\times n$ matrices). In the context of DNNs, the same design principle led to the Winograd filter-based convolution algorithm proposed by \citet{lavin2016fast}. This algorithm only requires $16$ instead of $36$ multiplications to compute $2\times2$ outputs of 2D convolutions with $3\times3$ kernels and achieves a $2$--$3\times$ speedup on GPU in practice.

From a hardware perspective, multipliers occupy considerably more area on chip than adders (for fixed-point data types). Field-programmable gate arrays (FPGAs) and application-specific integrated circuits (ASICs) can therefore potentially accommodate considerably more adders than multipliers, and trading off multiplications against additions is desirable. In fact, it was demonstrated recently that DNN architectures which rely on a large number of additions and a small number of multiplications (such as \cite{li2016ternary}) achieve a $60$\% higher throughput on FPGA than on GPU, while being $2.3\times$ better in performance per watt~\cite{nurvitadhi2017can}. In the context of ASICs, reducing the number of multiplications is beneficial as multiplication operations consume significantly more energy than addition operations ($3$--$30\times$ depending on the data type \cite{horowitz20141, andri2017yodann}). More generally, replacing multiplications in DNNs by additions leads to a reduction in models size as addition/subtraction can be encoded as a binary weight. This is beneficial in terms of throughput for most deep learning applications, which are typically memory-bound.

\newcommand{\picfont}{\small}
\begin{figure*}[ht!]
\vspace{-0.15cm}
\centering
\begin{tikzpicture}
    \node[anchor=south west,inner sep=0] (imagesp) at (0,0) {\includegraphics[width=0.25\textwidth]{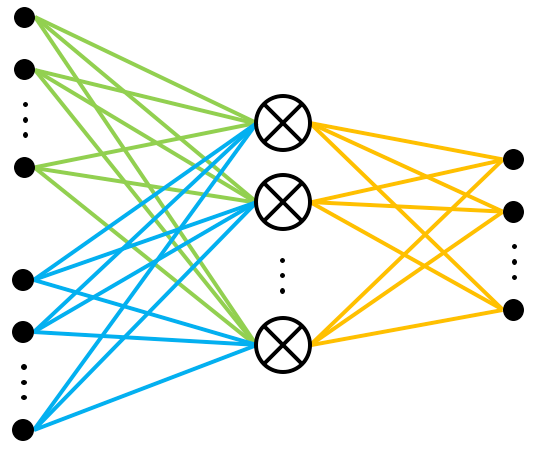}};
    \begin{scope}[x={(imagesp.south east)},y={(imagesp.north west)}]
        \node[] at (-0.13, 0.22) {\picfont$\vec(A)$};
        \node[] at (-0.13, 0.79) {\picfont$\vec(B)$};
        \node[] at (1.14, 0.48) {\picfont$\vec(C)$};
        
        \node[] at (0.32, 0.91) {\color{YellowGreen!80!white}\picfont$W_b$};
        \node[] at (0.8, 0.75) {\color{YellowOrange!80!white}\picfont$W_c$};
        \node[] at (0.32, 0.06) {\color{Cerulean!90!white}\picfont$W_a$};
    \end{scope}
\end{tikzpicture}
\quad
\begin{tikzpicture}
    \node[anchor=south west,inner sep=0] (image) at (0,0) {\includegraphics[width=0.55\textwidth]{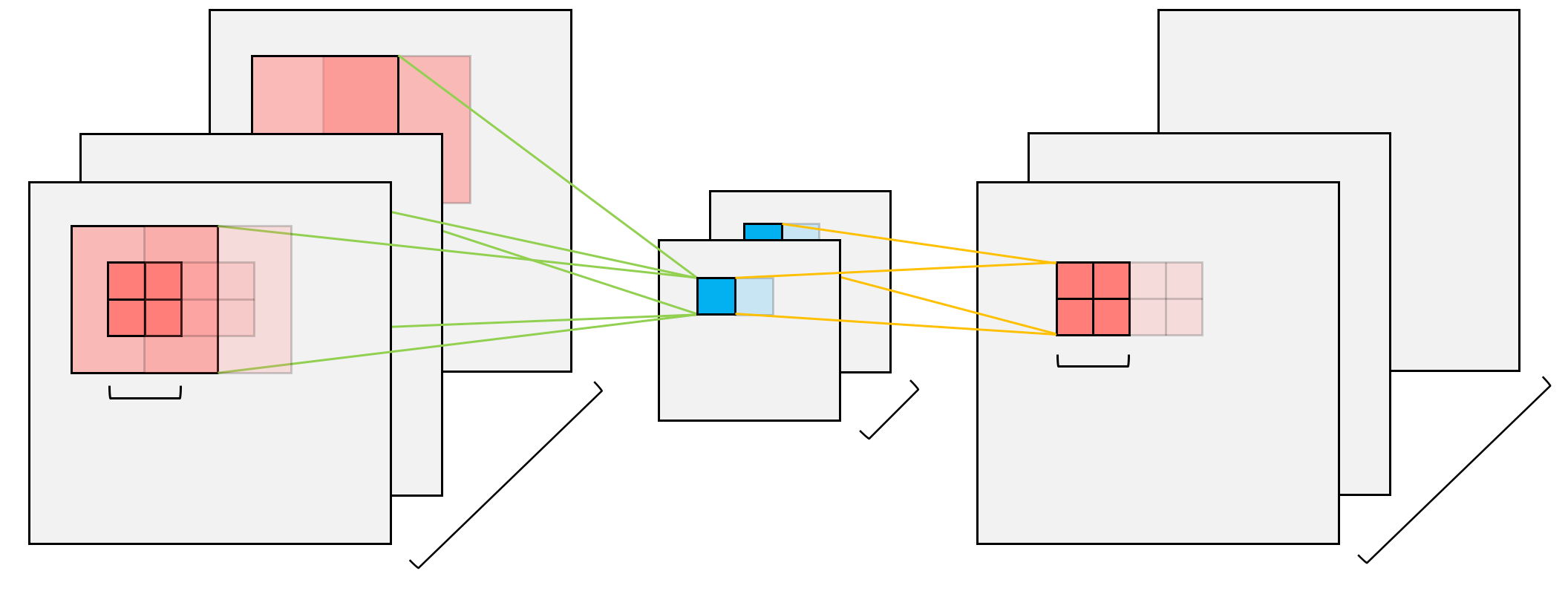}};
    \begin{scope}[x={(image.south east)},y={(image.north west)}]
        \node[] at (0.59, 0.28) {\picfont$r$};
        \node[] at (0.36, 0.19) {\picfont$\cin$};
        \node[] at (0.97, 0.19) {\picfont$\cout$};
        \node[] at (0.48, 0.23) {\picfont$\odot \tilde a$};
        \node[] at (0.09, 0.26) {\picfont$p$};
        \node[] at (0.695, 0.31) {\picfont$p$};
        
        \node[] at (0.41, 0.7) {\color{YellowGreen!80!white}\picfont$W_b$};
        \node[] at (0.595, 0.7) {\color{YellowOrange!80!white}\picfont$W_c$};
    \end{scope}
\end{tikzpicture}
\vspace{-0.45cm}
\caption{\label{fig:convillustration} {\bf Left:} Illustration of the 2-layer SPN \eqref{eq:strnet}, implementing an (approximate) matrix multiplication. The edges (i.e., the matrices $W_a$, $W_b$, $W_c$) have weights in $\K = \{-1,0,1\}$. {\bf Right:} Application of the proposed framework to 2D convolution leads to $p$-strided 2D convolution with $W_b$, followed by channel-wise scaling by $\tilde a = W_a \vec(A)$, followed by $1/p$-strided transposed 2D convolution with $W_c$.}
\vspace{-0.4cm}
\end{figure*}

Motivated by these observations, we propose a novel framework to drastically reduce the number of multiplications used by DNNs for inference.
Specifically, for every DNN layer, we cast the (matrix) multiplication of the weight matrix with the activations as a $2$-layer sum-product network (SPN) (arithmetic circuit). The SPNs disentangle (scalar) multiplications and additions in a way similar to Strassen's algorithm. The number of hidden units in the SPNs therefore determines the multiplication budget of the corresponding DNN layers. We then \emph{learn the addition and multiplication operations for all layers jointly from data} by learning the edges of the SPNs, encoded as ternary $\{-1,0,1\}$ matrices. As the transforms realized by the SPNs are approximate and adapted to the weight matrices and distribution of the activation tensors in the DNN, this allows us to reduce the number of multiplications much more drastically than hand-engineered transforms like Strassen's algorithm or the more specialized Winograd filter-based convolution. In summary, our main contributions are the following.
\vspace{-0.2cm}
\begin{itemize}
\setlength\itemsep{2pt}
\item We propose a SPN-based framework for stochastic gradient-based end-to-end learning of fast approximate transforms for the arithmetic operations in DNN layers. 
\item Our framework allows fine-grained control of the number of multiplications and additions used at inference time, enabling precise adjustment of the tradeoff between arithmetic complexity and accuracy of DNN models.
\item Extensive evaluations on CIFAR-10 and ImageNet show that our method applied to ResNet \cite{he2016deep} yields the same or higher accuracy than existing complexity reduction methods while using considerably fewer multiplications. For example, for ResNet-18 our method reduces the number of multiplications by 99.63\% while incurring a top-1 accuracy degradation of only 2.0\% compared to the full-precision model on ImageNet.
\item Our method applied to a language model with convolution and LSTM layers \cite{kim2016character} results in a 99.69\% reduction in multiplications while inducing an increase of only 3.3\% in perplexity.

\item Combining our method with knowledge distillation (KD) techniques, we obtain for the first time massive reductions in number of multiplications (99.5\% and more) while maintaining the predictive performance of the full-precision models, for both image classification and language modeling.

\item We demonstrate that the proposed framework is able to rediscover Strassen's algorithm, i.e., it can learn to (exactly) multiply $2 \times 2$ matrices using only $7$ multiplications instead of $8$.
\end{itemize}
\vspace{-0.2cm}

Two key aspects of our approach that lead to gains compared previous methods are (i) our method is specifically tailored to reduce the number of multiplications whereas some previous works put more emphasis on model size reduction, and (ii) we leverage knowledge distillation which improves our results further.

We continue by reviewing related work in Section \ref{sec:relwork} and then describe our method and its application to 2D convolution in Section \ref{sec:concept}. A detailed numerical evaluation of our method is presented in Section \ref{sec:experiments} and concluding remarks can be found in Section \ref{sec:conclusion}. 

\subsection{Related work} \label{sec:relwork}

We briefly review the most common approaches to compress DNNs, focusing on methods decreasing computational complexity rather than memory footprint. In all cases, there is a tradeoff between the complexity reduction and reduction in the (inference) accuracy of the compressed model.

A popular way to speed up DNNs, in particular convolutional neural networks (CNNs), is to utilize resource-efficient architectures, such as SqueezeNet \cite{iandola2016squeezenet}, MobileNet \cite{howard2017mobilenets}, and ShuffleNet \cite{zhang2017shufflenet}. SqueezeNet reduces the convolution kernel size. MobileNet and ShuffleNet rely on depth-wise separable convolutions and grouped convolutions, respectively. More sophisticated grouping and sharing techniques are studied by \citet{wangdesign}.

Another strategy to accelerate CNNs is to exploit the low-rank structure prevalent in weight matrices and convolution kernels. \citet{denton2014exploiting, novikov2015tensorizing, kim2015compression} use tensor decompositions to obtain low-rank approximations of pretrained weight matrices and filter tensors, then finetune the approximated weight matrices and filters to restore the accuracy of the compressed models. Other works \cite{tai2015convolutional, wen2017coordinating} employ low rank-promoting regularizers to further reduce the rank of the filter tensors. A framework to exploit low-rank structure in the filter responses is presented by \citet{zhang2016accelerating}.

Sparsifying filters and pruning channels are popular methods to make DNNs more efficient during inference. \citet{wen2016learning} and \citet{lebedev2016fast} rely on group norm-based regularizers and demonstrate their effectiveness in penalizing unimportant filters and channels, promoting hardware-friendly filter shapes, regularizing the network depth, and optimizing the filter receptive fields. Inter-channel and intra-channel redundancy is exploited by \citet{liu2015sparse} via a two-stage factorization procedure. An energy-aware methodology to prune filters of CNNs is described in \cite{yang2016designing}.

Finally, an effective way to adapt DNNs to resource-constrained platforms is to reduce the numerical precision of their weights and/or activations. Examples for DNNs that quantize both weights and activations are DoReFa-Net \cite{zhou2016dorefa}, XNOR-Net \cite{rastegari2016xnor}, and ABC-Net \cite{lin2017towards}. Other works use binary weights \cite{courbariaux2015binaryconnect, rastegari2016xnor, lin2017towards} and ternary weights \cite{li2016ternary, zhu2016trained} but maintain full-precision values for the activations. Keeping the activations in full precision instead of quantizing them leads to a smaller decrease in computational cost, but can yield better predictive performance.

\section{Learning Fast Matrix Multiplications via SPNs}

\label{sec:concept}

\subsection{Casting matrix multiplication as SPN}

Given square matrices $A, B \in \R^{n \times n}$, the product $C = A B$ can be represented as a $2$-layer SPN
\begin{equation}
    \label{eq:strnet}
    \vec(C) = W_c [ (W_b \vec(B)) \odot (W_a \vec(A))]
\end{equation}
where $W_a, W_b \in \K^{r \times n^2}$ and $W_c \in \K^{n^2 \times r}$, with $\K \coloneqq \{-1,0,1\}$, are fixed, $\vec(D)$ stands for vectorization of the matrix $D=[d_1 \, \ldots \, d_m] \in \R^{k \times m}$, i.e., $\vec(D) = [d_1^\top \, \ldots \, d_m^\top]^\top \in \R^{k m}$, and $\odot$ denotes the element-wise product. The SPN \eqref{eq:strnet} disentangles additions (and subtractions), encoded in the ternary matrices $W_a$, $W_b$, and $W_c$, and multiplications, realized exclusively by the operation $\odot$ (see Fig. \ref{fig:convillustration}, left). The width of the hidden layer of the SPN, $r$, hence determines the number of multiplications used for the matrix multiplication. A na{\"i}ve implementation of the matrix multiplication $AB$ requires $r = n^3$. For $n=2$,\footnote{The formulation by \citet{strassen1969gaussian} is more general, applying recursively to $4$ equally-sized subblocks of square matrices, with the $2\times2$ case occurring at maximal recursion depth.} Strassen's matrix multiplication algorithm~\cite{strassen1969gaussian} specifies the following set of weights that satisfy \eqref{eq:strnet} for $r=7$ (instead of $r = 8$)
\begin{align}
&\begingroup
\setlength\arraycolsep{3.5pt}
W_a = 
\begin{pmatrix*}[r]
1 & 0 & 0 & 1 \\
0 & 1 & 0 & 1 \\
1 & 0 & 0 & 0 \\
0 & 0 & 0 & 1 \\
1 & 0 & 1 & 0 \\
-1 & 1 & 0 & 0 \\
0 & 0 & 1 & -1
\end{pmatrix*},
\; 
W_b = 
\begin{pmatrix*}[r]
1 & 0 & 0 & 1 \\
1 & 0 & 0 & 0 \\
0 & 0 & 1 & -1 \\
-1 & 1 & 0 & 0 \\
0 & 0 & 0 & 1 \\
1 & 0 & 1 & 0 \\
0 & 1 & 0 & 1
\end{pmatrix*},
\endgroup
\nonumber \\
&\qquad \quad\begingroup
\setlength\arraycolsep{3.5pt}
W_c = 
\begin{pmatrix*}[r]
1 & 0 & 0 & 1 & -1 & 0 & 1 \\
0 &1 & 0 & 1 & 0 & 0 & 0 \\
0 & 0 & 1 & 0 & 1 & 0 & 0 \\
1 & -1 & 1 & 0 & 0 & 1 & 0
\end{pmatrix*}. 
\endgroup \label{eq:strassen}
\end{align}
An interesting tensor perspective on the SPN \eqref{eq:strnet} (not explored in-depth here) is common in the context of algebraic complexity theory. Specifically, \eqref{eq:strnet} can be written as
\begin{align*}
&\vec(C)_i = \sum_{k=1}^{n^2} \sum_{l=1}^{n^2} (M_n)_{i,k,l} \vec(A)_k \vec(B)_l, \; \text{where} \\ & \qquad (M_n)_{i,k,l} = \sum_{j=1}^r (W_c)_{i,j} (W_a)_{j,k} (W_b)_{j,l}.
\end{align*}
$M_n$ is the \emph{$(n \times n)$-matrix multiplication tensor}, and $r$ hence corresponds to the rank of $M_n$. It is known that $\rank(M_2) = 7$ and $19 \leq \rank(M_3) \leq 23$, see \cite{elser2016network} for more details and references. 

\citet{elser2016network} explores learning exact matrix multiplications via SPNs of the form \eqref{eq:strnet} for $n=2$ and $n=3$ from synthetic data. Thereby, the elements of $W_a$, $W_b$, and $W_c$ are relaxed to real numbers instead of elements from $\K$. Note that this relaxation leads to an increase in the number of multiplications in general. In contrast, we integrate SPNs with weights from $\K$ into DNN layers and learn them end-to-end (see next section), realizing actual reductions in multiplications.

\subsection{Learning fast approximate matrix multiplications for DNNs} \label{sec:learning}

Writing matrix products in the form \eqref{eq:strnet} is not specific to square matrices. Indeed, it is easy to see that $r \geq nmk$ is a sufficient condition for the existence of matrices $W_a, W_b, W_c$ with elements in $\K$ such that the product of any two matrices $A \in \R^{k \times m}$ and $B \in \R^{m \times n}$, including matrix-vector products (i.e., $n=1$), can be written in the form \eqref{eq:strnet}. When the matrices $A$ and $B$ are drawn from probability distributions that concentrate on low-dimensional manifolds of $\R^{k \times m}$ and $\R^{m \times n}$, respectively, or if one of the matrices is fixed, it may be possible to find $W_a$ and $W_b$ that satisfy the equality in \eqref{eq:strnet} approximately even when $r \ll nmk$. In this case, \eqref{eq:strnet} approximately computes the product $AB$ while considerably reducing the number of multiplications compared to the na{\"i}ve implementation. Furthermore, by imposing structure (such as, e.g., sparsity or block-diagonal structure) into the matrices $W_a$, $W_b$, $W_c$ one can tailor sharing or grouping of the operations for the application or platform at hand.  

In this paper, we leverage this concept to accelerate and compress the matrix multiplications in DNN layers for inference. Specifically, for layer $\ell$, we associate $A$ with the (pretrained) weights/filters $W_\ell$ and $B$ with the corresponding activations/feature maps $F_\ell$. The ternary matrices $W_a$, $W_b$, and $W_c$ are then learned end-to-end using a stochastic gradient-based optimizer (one set of weights $W_a$, $W_b$, $W_c$ for each layer). 
After training, $W_a$ and $\vec(A)$ can be collapsed into a vector $\tilde a = W_a \vec(A) \in \R^r$ as they are both fixed during inference. Alternatively, $\tilde a \in \R^r$, $W_b$, and $W_c$ can be learned jointly from scratch. The choice of $r$ determines the tradeoff between the computational cost in terms of multiplications and the precision of the the approximate matrix multiplication, and hence the predictive performance of the network. This approach requires $r$ full-precision parameters and $rm(k+n)$ ternary weight parameters. It reduces the number of multiplications by a factor of $mnk/r$.

Quantizing the elements of $W_a$, $W_b$, and $W_c$ to $\K$ during training poses a challenge as quantization is non-differentiable. Different approaches were proposed to overcome this issue \cite{courbariaux2015binaryconnect, li2016ternary, rastegari2016xnor, zhu2016trained, agustsson2017soft}. Here, we adopt the method from \cite{li2016ternary} and briefly describe it for quantizing $W_a$ ($W_b$ and $W_c$ are quantized in exactly the same way). Specifically, this method maintains a full-precision version $W_a^\mathrm{fp}$ of $W_a$ during training and quantizes $W_a^\mathrm{fp}$ in every forward pass by approximately solving the optimization problem
\begin{align}\label{eq:quantprob}
 \alpha^*, W_a^\mathrm{t*} &= \underset{\alpha, W_a^\mathrm{t}}{\arg\min} \| W_a^\mathrm{fp} - \alpha W_a^\mathrm{t} \|_F^2 \nonumber \\
 & \qquad \text{s.t.} \quad \alpha>0, \quad
 W_a^\mathrm{t} \in \K^{r \times km},
\end{align}
and by setting $W_a = \alpha^* W_a^\mathrm{t*}$ (the scaling factors $\alpha^*$ for $W_a$, $W_b$, $W_c$ can be absorbed by $A$ or $\tilde a$ after training to ensure that $W_a$, $W_b$, $W_c$ have elements in $\K$). During the backward pass the quantization function is replaced by the identity function, and the gradient step is applied to $W_a^\mathrm{fp}$. 
Assuming i.i.d. Gaussian weights, \citet{li2016ternary} derive the approximate solution
\begin{align}\label{eq:quantapprox}
(W_a^\mathrm{t*})_{i,j} &= 
\begin{cases}
\hphantom{-}1 \quad&\text{if} \; (W_a^\mathrm{fp})_{i,j} > \Delta, \\
-1 &\text{if} \; (W_a^\mathrm{fp})_{i,j} < -\Delta, \\
\hphantom{-}0 &\text{otherwise},
\end{cases} \nonumber \\
\alpha^* &= \frac{\sum_{(i,j) \colon (W_a^\mathrm{t*})_{i,j} \neq 0} |(W_a^\mathrm{fp})_{i,j}|}{\sum_{i,j}  |(W_a^\mathrm{t*})_{i,j}|}
\end{align}
 to \eqref{eq:quantprob}, where $\Delta = \frac{0.7}{kmr} \sum_{i,j}  |(W_a^\mathrm{fp})_{i,j}|$. 
While our framework would allow quantized training from scratch with fixed threshold $\Delta$ and fixed quantization level $\alpha$ (e.g., $\Delta=0.5$ and $\alpha=1$), we observed that relying on the scheme \eqref{eq:quantapprox} allows us to pretrain $W_a^\mathrm{fp}$, $W_b^\mathrm{fp}$, $W_c^\mathrm{fp}$ without quantization, and then activate quantization to stably continue training. We found that this strategy leads to faster training while inducing no loss in accuracy.

Besides the fully connected case described in this section, we particularize the proposed approach for 2D convolutions for image classification DNNs. We emphasize that  any DNN layer operation reducible to a general matrix multiplication (GEMM) can be cast into the form \eqref{eq:strnet}, including $n$-dimensional convolutions, group (equivariant) convolutions (when implemented as a filter bank) \cite{cohen2016group}, and deformable convolutions \cite{dai2017deformable}.

\subsection{Knowledge distillation (KD)}

KD refers to the process of training a student network using a larger (in terms of the number of layers and hidden units) teacher network \cite{bucilua2006model, hinton2015distilling}. As a result, the student network typically has the same or slightly better predictive performance than the teacher network, despite being less complex. KD for training a low-precision student network from a full-precision teacher network with the same architecture and hyper parameters as the student network was investigated recently in \cite{mishra2017apprentice, zhuang2017towards, polino2018model}. Here, we explore the same avenue to improve the predictive performance of networks compressed with our method. Specifically, we follow the method proposed in \cite{hinton2015distilling} using the cross entropy between the student softmax output and the teacher softmax output as KD loss term. We set the softmax temperature parameter to $1$ throughout and assign the same weight to the KD loss term as to the original loss. For sequence models, we simply apply the described KD loss to the softmax outputs of the unrolled teacher and student models (more sophisticated techniques were proposed in \cite{kim2016sequence}).

\subsection{Application to 2D convolution} \label{sec:2dconv}

Consider the $\ell$th 2D convolution layer of a CNN applying $\cout$ filters of dimension $w \times h \times \cin$ 
to a feature representation $F_\ell$ of dimension $W \times H \times \cin$ (width$\times$height$\times$number of channels). To write the computation of all $\cout$ output channels as a matrix multiplication, each feature map in $F_\ell$ is decomposed into $WH$ patches of size $w \times h$ (after appropriate padding) and the vectorized patches are arranged in a matrix $\tilde F_\ell$ of dimension $w h \cin \times WH$. This transformation is usually referred to as {\tt im2col}, see (\citet{sze2017efficient}, Fig. 19) for an illustration. Accordingly, the filters for all output channels are vectorized and jointly reshaped into a $\cout \times w h \cin$ matrix $\tilde W_\ell$. The vectorized layer output (before activation) for all $\cout$ output channels is obtained as $\tilde W_\ell \tilde F_\ell$ and has dimension $\cout \times WH$. In principle, one can now compress the operation $\tilde W_\ell \tilde F_\ell$ using our method by setting $A = \tilde W_\ell$,  $B = \tilde F_\ell$, plugging them into \eqref{eq:strnet}, and proceeding as described in Section~\ref{sec:learning}. However, this results in impractically large $W_a$, $W_b$, and $W_c$ and ignores the weight sharing structure of the convolution. By associating $A$ with $\tilde W_\ell$ and $B$ with single columns of $\tilde F_\ell$ we can jointly compress the computations across all input and output channels, while preserving the spatial structure of the convolution. The resulting SPN realizes a convolution with $r$ ternary $w \times h \times \cin$ filters (the rows of $W_b$), followed by a channel-wise scaling with $\tilde a = W_a\vec(\tilde W_\ell)$, followed by convolution with a ternary $1 \times 1 \times r$ filter for each of the $\cout$ outputs (the rows of $W_c$) see Fig. \ref{fig:convillustration}, right. 

To realize \emph{local spatial compression}, we partition the computation of the convolution into subsets corresponding to square output patches. In more detail, we consider the computation of $p \times p$ convolution output patches from $(p - 1 + w) \times (p - 1 + h)$ input patches, offset by a stride of $p$, and approximate this computation with a SPN jointly for all channels. As a result, the number of multiplications is reduced both spatially and across channels. For example, for $3 \times 3$ convolution filters,
we divide the input feature maps into $4\times4$ spatial patches with a stride of $2$, 
such that the SPN computes $2 \times 2 \times \cout$ outputs from $4 \times 4 \times \cin$ elements of $F_\ell$. Thereby, $W_c$ realizes a $2 \times 2 \times r$ transposed convolution with a stride of $1/2$ 
(see Fig. \ref{fig:convillustration}, right, and pseudocode in Appendix~\ref{sec:2dconvpseudo}). 
For fixed $r$, this reduces the number of multiplications by a factor of $4$ compared to the case without spatial compression (i.e., $p=1$). 

In summary, the described compression of 2D convolution 
leads to a reduction of the number of multiplications 
by a factor $\cin\cout w h p^2 /r$ 
compared to the standard implementation of the convolution. 

Finally, to reduce the number of additions realized through $W_b$ (and thereby the number of nonzero elements of $W_b$) by a factor of $g$, we implement $W_b$ as grouped convolution, originally introduced in \cite{krizhevsky2012imagenet}. Specifically, the convolution realized by $W_b$ is assumed to consist of $g$ independent 2D convolutions each with $\cin/g$ input channels and $r/g$ output channels. In other words, $W_b$ is assumed to be block-diagonal with blocks of dimension $(r/g)\times(wh\cin/g)$.

{\bf Relation to prior work in the 2D convolution case.}
Binary weight networks (BWNs) \cite{rastegari2016xnor} and ternary weight networks (TWNs) \cite{li2016ternary} rely on binary $\{-1,1\}$ and ternary $\{-1,0,1\}$ weight matrices, respectively, followed by (full-precision) rescaling of the activations (see Section \ref{sec:learning}) and are special cases of our framework. ABC-Nets \cite{lin2017towards} approximate the full-precision weight matrices as a weighted sum of multiple binary $\{-1,1\}$ weight matrices and can also be cast as (structured) SPNs. However, we do not directly recover the trained ternary quantization (TTQ) approach from \cite{zhu2016trained}, which relies on asymmetric ternary weights $\{-c_1, 0, c_2\}$, $c_1,c_2 >0$. Finally, note that Winograd filter-based convolution \cite{lavin2016fast} realizes spatial compression over $2\times2$ output patches but performs exact computation and does not compress across channels.

\section[Experiments]{Experiments\footnote{Code available at {\url{https://github.com/mitscha/strassennets}}.}} \label{sec:experiments}

\subsection{Rediscovering Strassen's algorithm}

Before applying the proposed method to DNNs, we demonstrate that it is able to rediscover Strassen's algorithm, i.e., it can learn to multiply $2 \times 2$ matrices using only $7$ multiplications instead of $8$ (which implies a recursive algorithm for larger matrices). This problem was previously studied by \citet{elser2016network}, but for \emph{real-valued} $W_a$, $W_b$, $W_c$, which increases the number of multiplications in general when using these matrices in \eqref{eq:strnet} to compute matrix products. 
In contrast, our method learns $W_a, W_b \in \K^{7 \times 4}$, $W_c\in \K^{4 \times 7}$ (i.e., the discrete solution space has size $3^{3\cdot4\cdot7}=3^{84}$), and hence leads to an actual reduction in the number of multiplications.

We generate a training set containing 100k pairs $(A_i, B_i)$ with entries i.i.d. uniform on $[-1,1]$, train the SPN with full-precision weights (initialized i.i.d. uniform on $[-1,1]$) for one epoch with SGD (learning rate $0.1$, momentum $0.9$, mini-batch size $4$), activate quantization, and train for another epoch (with learning rate $0.001$). Around $25$ random initializations are necessary to obtain convergence to zero training L2-loss after activation of the quantization; for most initializations the training L2-loss converges to a positive value. A set of ternary weight matrices implementing an \emph{exact} matrix multiplication, found by our method, is
\begin{align}
&\begingroup
\setlength\arraycolsep{3.5pt}
W_a = 
\begin{pmatrix*}[r]
-1 & -1 & 0 & 0 \\
0 & 0 & 0 & 1 \\
-1 & -1 & 1 & 1 \\
-1 & 0 & 1 & 0 \\
-1 & -1 & 1 & 0 \\
0 & 0 & 1 & 0 \\
0 & -1 & 0 & 0
\end{pmatrix*},
\; 
W_b = 
\begin{pmatrix*}[r]
-1 & -1 & 0 & 0 \\
0 & 0 & 0 & 1 \\
0 & 1 & 0 & 0 \\
1 & 0 & 1 & 0 \\
-1 & -1 & -1 & 0 \\
1 & 1 & 1 & 1 \\
0 & 0 & -1 & 0
\end{pmatrix*},
\endgroup \nonumber \\
&\qquad \quad \begingroup
\setlength\arraycolsep{3.5pt}
W_c = 
\begin{pmatrix*}[r]
1 & 0 & 0 & -1 & -1 & 0 & 1 \\
0 & 0 & 1 & 1 & 1 & 0 & -1 \\
-1 & 0 & 0 & 0 & 1 & 1 & -1 \\
0 & 1 & 0 & 0 & 0 & 0 & 1
\end{pmatrix*}. 
\endgroup \nonumber
\end{align}
\vspace{-0.4cm}

\subsection{Image classification}

We apply our method to all convolution layers (including the first convolution layer and the projection layers for subsampling) of the ResNet architecture \cite{he2016deep} to create the so-called Strassen-ResNet (ST-ResNet). We evaluate ST-ResNet on CIFAR-10 (10 classes, 50k training images, 10k testing images) \cite{krizhevsky2009learning} and ImageNet (ILSVRC2012; 1k classes, 1.2M training images, 50k testing images) \cite{ILSVRC15} for different choices of $r$, $p$, $g$, and compare the accuracy of ST-ResNet to related works. All models were trained from scratch, meaning we directly learn $\tilde a = W_a \vec(A)$ rather than associating $A$ with the weights of pretrained networks and learning $W_a$. Throughout the training process we used SGD with momentum $0.9$ and weight decay $10^{-4}$. 
As most related works involving ternary weights do not report sparsity levels, to facilitate comparisons, we do not make any assumption about the number of zeros among ternary weights. It is the sparsity of the activations, not the weights, that directly impacts the number of multiplications (the focus of this paper). 
All model sizes are computed without (lossless) compression of the network parameters. 

\subsubsection[CIFAR-10]{CIFAR-10}  \label{sec:cifar}

We consider ST-ResNet-20 and employ the data augmentation procedure described in (\citet{he2016deep}, Sec. 4.2.). We train for $250$ epochs with initial learning rate $0.1$ and mini-batch size $128$, multiplying the learning rate by $0.1$ after $150$ and $200$ epochs. We then activate quantization for $W_b$ and $W_c$, set the learning rate to $0.01$ and train the network for $40$ epochs, multiplying the learning rate by $0.1$ every 10 epochs. Finally, we fix the (now ternary) $W_b$ and $W_c$ and continue training for another $10$ epochs. The resulting testing accuracy is shown in Table \ref{tab:acccifar} for different $r$ and $p$, along with the reduction in the number of multiplications compared to the uncompressed model (for the $32 \times 32$ CIFAR-10 images; see Table \ref{tab:compcifarapp} in Appendix \ref{sec:addres} for the reduction in the number of additions). Additional results for a similar experiment based on the VGG-inspired 7-layer architecture considered in \cite{courbariaux2015binaryconnect, li2016ternary} can be found in Appendix~\ref{sec:vggcifar}.

\setlength{\extrarowheight}{1mm}

\vspace{-0.3cm}\begin{table}[h!]
  \caption{\label{tab:acccifar}{\bf Left:} Testing accuracy (in \%) of compressed ResNet-20 on CIFAR-10. {\bf Right}: Reduction in the number of multiplications.}
  \centering
  \small
  \setlength{\tabcolsep}{0.9mm}
  \begin{tabular}{lccccc}
  \multicolumn{5}{c}{testing accuracy} \\
    \toprule
    & \multicolumn{4}{c}{$r$} \\ \cline{2-5}
$p$ & $\cout$ & $\frac{3}{4}\cout$ & $\frac{1}{2}\cout$ & $\frac{1}{4}\cout$ \\[1mm]
\hline
$1$ & 91.24 & 90.62 & 88.63 & 85.46 \\
$2$ & 89.87 & 89.47 & 87.31 & 84.01  \\
$4$ & 86.13 & 84.67 & 82.67 & 75.01 \\
    \bottomrule
  \end{tabular}
  \,
  \begin{tabular}{lcccc}
    \multicolumn{5}{c}{red. in multiplications} \\
\toprule
& \multicolumn{4}{c}{$r$} \\ \cline{2-5}
$p$ & $\cout$ & $\frac{3}{4}\cout$ & $\frac{1}{2}\cout$ & $\frac{1}{4}\cout$ \\[1mm]
\hline
$1$ & 98.96 & 99.08 & 99.21 & 99.33 \\
$2$ & 99.33 & 99.36 & 99.39 & 99.42 \\
$4$ & 99.42 & 99.43 & 99.44 & 99.44 \\
  \bottomrule
  \end{tabular}
\end{table}

{\bf Discussion.} The model obtained for the base configuration with $r = \cout$ and $p = 1$  incurs a negligible accuracy loss compared to the uncompressed ResNet-20 with an accuracy of $91.25\%$ \cite{he2016deep} while reducing the number of multiplications by $98.96\%$ (the evaluation of the uncompressed ResNet-20 requires $41.038$M multiply-adds). This model also matches the accuracy of TTQ \cite{zhu2016trained} for ResNet-20 while requiring fewer multiplications (TTQ does not quantize the first convolution layer). As $r$ decreases and/or $p$ increases, the number of multiplications decreases at the cost of further accuracy reduction.

\subsubsection{ImageNet}

We consider ST-ResNet-18 and, unlike for the experiment on CIFAR-10, we also compress the last (fully connected) layer of ST-ResNet-18 for models with $r \leq \cout$ in convolution layers, setting $r=1000$ for that layer throughout (we observed that compressing the last layer when $r > \cout$ in convolution layers leads to a considerable reduction in validation accuracy). Following  \cite{rastegari2016xnor, li2016ternary, zhu2016trained}, the training images are resized such that the shorter side has length $256$ and are then randomly cropped to $224 \times 224$ pixels. The validation accuracy is computed from center crops. We use an initial learning rate of $0.05$ and mini-batch size $256$, with two different learning rate schedules depending on the value of $r$ in the convolution layers: We train for 40 epochs without quantization, multiplying the learning rate by $0.1$ after 30 epochs, if $r \leq \cout$, and for 70 epochs, multiplying the learning rate by $0.1$ after $40$ and $60$ epochs, otherwise. Thereafter, we activate quantization and continue training for 10 epochs. Finally, we fix $W_b$ and $W_c$ and train $\tilde a$ for another 5 epochs. 

In Table \ref{tab:accimg} we report the validation accuracy of ST-ResNet-18 for different $r$, $p$, and $g$, and the validation accuracy obtained with KD. Table \ref{tab:redimg} shows the reduction in the number of multiplications compared to the original ResNet-18 model, for different $r$, $p$, and $g$ (see Table \ref{tab:redimgapp} in Appendix~\ref{sec:addres} for reductions in the number of additions and model size). 
In Fig. \ref{fig:curves}, we plot the accuracy of ST-ResNet-18 for different $r$, $p$, and $g$, as a function of the number of operations and model size. In addition, we report the validation accuracy for related works \cite{rastegari2016xnor, li2016ternary, zhu2016trained, lin2017towards} (see also Table \ref{tab:accrelimg} in  Appendix~\ref{sec:addres}). We do not consider $p > 2$ as this leads to (ternary) convolution with impractically large kernels for $224\times224$ images.

Finally, to demonstrate amenability of our method to larger models, we trained ST-ResNet-34 with $r=2\cout$, $p=2$, $g=1$ (without tuning any hyper parameters) and obtained 69.2\%/88.5\% top-1/top-5 validation accuracy without KD and 71.9\%/90.5\% with KD (the full-precision model obtains 73.3\%/91.3\%; we report the accuracies of the Torch pretrained models for all full-precision ResNets).

\newcommand{\msimg}{1.3pt}

\newcommand{\colora}{green!65!black}
\newcommand{\colorb}{blue}
\newcommand{\colorc}{red}
\newcommand{\legsize}{\tiny}

\begin{figure}[t]
\centering
\hspace{-0.13cm}\begin{tikzpicture}[scale=1, every node/.style={scale=0.9}] 
\begin{groupplot}[group style={group size=3 by 2, vertical sep=0.1cm, horizontal sep=0.1cm},height=5.1cm,width=4cm,
enlargelimits=0.1, grid=major, /tikz/font=\small] 
    
\nextgroupplot[xmode=log, ylabel={top-1 acc. [\%]},
        nodes near coords align=right, xmin=3.5e6, xmax=3e9,
        y label style={at={(axis description cs:0.2,.5)}},
	xticklabels={,,},] \addplot[scatter, color=\colora,
	scatter/classes={ 
	0={mark=triangle, \colora},
	1={mark= square,\colora},
	2={mark=o, \colora},
	3={mark=diamond, \colora},
	4={mark=pentagon, \colora}
	}, 
	scatter src=explicit symbolic, mark size=\msimg,]
	table[select coords between index={0}{4},x index=0,y index=1, meta=class] \toponemul;

\addplot[scatter, color=\colora,
	scatter/classes={ 
	0={mark=square*, \colora},
	1={mark=*,\colora}
	}, 
	scatter src=explicit symbolic, mark size=\msimg, only marks]
	table[select coords between index={0}{0},x index=0,y index=1, meta=class] \toponemuldist;
	
\addplot[scatter, color=\colorb,
	scatter/classes={ 
	0={mark=triangle,\colorb},
	1={mark=square, \colorb},
	2={mark=o, \colorb},
	3={mark=diamond, \colorb},
	4={mark=pentagon, \colorb}
	}, 
	scatter src=explicit symbolic, mark size=\msimg,]
	table[select coords between index={5}{9},x index=0,y index=1, meta=class] \toponemul;

\addplot[scatter, color=\colorb,
	scatter/classes={ 
	0={mark=square*, \colorb},
	1={mark=*,\colorb}
	}, 
	scatter src=explicit symbolic, mark size=\msimg, only marks]
	table[select coords between index={1}{2},x index=0,y index=1, meta=class] \toponemuldist;
	
\addplot[scatter, color=\colorc,
	scatter/classes={ 
	0={mark=triangle, \colorc},
	1={mark=square, \colorc},
	2={mark=o, \colorc},
	3={mark=diamond, \colorc},
	4={mark=pentagon, \colorc}
	}, 
	scatter src=explicit symbolic, mark size=\msimg,]
	table[select coords between index={10}{14},x index=0,y index=1, meta=class] \toponemul;

\addplot[scatter, color=\colorc,
	font=\legsize,
	nodes near coords,
	only marks,
	mark=x,
	black,
	scatter src=explicit symbolic,]
	table[select coords between index={15}{18},x index=0,y index=1, meta=class] \toponemul;

\addplot[scatter, color=black,
	font=\legsize,
	nodes near coords,
	only marks,
	mark=+,
	scatter src=explicit symbolic,]
	table[select coords between index={19}{22},x index=0,y index=1] \toponemul;
	
\addplot[dashed, gray] coordinates {(2e6,69.6) (1e10,69.6)};
	
\nextgroupplot[xmode=log, nodes near coords align=below right, xmin=4.2e8, xmax=1.3e10,
xticklabels={,,},
yticklabels={,,},] \addplot[scatter, color=\colora,
	scatter/classes={ 
	0={mark=triangle, \colora},
	1={mark= square,\colora},
	2={mark=o, \colora},
	3={mark=diamond, \colora},
	4={mark=pentagon, \colora}
	},  
	scatter src=explicit symbolic, mark size=\msimg,]
	table[select coords between index={0}{4},x index=0,y index=1, meta=class] \toponeadd;

\addplot[scatter, color=\colora,
	scatter/classes={ 
	0={mark=square*, \colora},
	1={mark=*,\colora}
	}, 
	scatter src=explicit symbolic, mark size=\msimg, only marks]
	table[select coords between index={0}{0},x index=0,y index=1, meta=class] \toponeadddist;
	
\addplot[scatter, color=\colorb,
	scatter/classes={ 
	0={mark=triangle,\colorb},
	1={mark=square, \colorb},
	2={mark=o, \colorb},
	3={mark=diamond, \colorb},
	4={mark=pentagon, \colorb}
	}, 
	scatter src=explicit symbolic, mark size=\msimg,]
	table[select coords between index={5}{9},x index=0,y index=1, meta=class] \toponeadd;
	
\addplot[scatter, color=\colorb,
	scatter/classes={ 
	0={mark=square*, \colorb},
	1={mark=*,\colorb}
	}, 
	scatter src=explicit symbolic, mark size=\msimg, only marks]
	table[select coords between index={1}{2},x index=0,y index=1, meta=class] \toponeadddist;	
	
\addplot[scatter, color=\colorc,
	scatter/classes={ 
	0={mark=triangle, \colorc},
	1={mark=square, \colorc},
	2={mark=o, \colorc},
	3={mark=diamond, \colorc},
	4={mark=pentagon, \colorc}
	}, 
	scatter src=explicit symbolic, mark size=\msimg,]
	table[select coords between index={10}{14},x index=0,y index=1, meta=class] \toponeadd;
	
\addplot[scatter, color=\colorc,
	font=\legsize,
	nodes near coords,
	only marks,
	mark=x,
	black,
	scatter src=explicit symbolic,]
	table[select coords between index={15}{18},x index=0,y index=1, meta=class] \toponeadd;
	
\addplot[scatter, color=black,
	font=\legsize,
	nodes near coords,
	only marks,
	mark=+,
	scatter src=explicit symbolic,]
	table[select coords between index={19}{22},x index=0,y index=1] \toponeadd;

\addplot[dashed, gray] coordinates {(3e8,69.6) (2e10,69.6)};

\nextgroupplot[xmode=log, nodes near coords align=below left, xmin=7,xmax=400,
xticklabels={,,},
yticklabels={,,},] \addplot[scatter, color=\colora,
	scatter/classes={ 
	0={mark=triangle, \colora},
	1={mark= square,\colora},
	2={mark=o, \colora},
	3={mark=diamond, \colora},
	4={mark=pentagon, \colora}
	},  
	scatter src=explicit symbolic, mark size=\msimg,]
	table[select coords between index={0}{4},x index=0,y index=1, meta=class] \toponespace;
	
\addplot[scatter, color=\colora,
	scatter/classes={ 
	0={mark=square*, \colora},
	1={mark=*,\colora}
	}, 
	scatter src=explicit symbolic, mark size=\msimg, only marks]
	table[select coords between index={0}{0},x index=0,y index=1, meta=class] \toponespacedist;
	
\addplot[scatter, color=\colorb,
	scatter/classes={ 
	0={mark=triangle,\colorb},
	1={mark=square, \colorb},
	2={mark=o, \colorb},
	3={mark=diamond, \colorb},
	4={mark=pentagon, \colorb}
	},
	scatter src=explicit symbolic, mark size=\msimg,]
	table[select coords between index={5}{9},x index=0,y index=1, meta=class] \toponespace;

\addplot[scatter, color=\colorb,
	scatter/classes={ 
	0={mark=square*, \colorb},
	1={mark=*,\colorb}
	}, 
	scatter src=explicit symbolic, mark size=\msimg, only marks]
	table[select coords between index={1}{2},x index=0,y index=1, meta=class] \toponespacedist;
	
\addplot[scatter, color=\colorc,
	scatter/classes={ 
	0={mark=triangle, \colorc},
	1={mark=square, \colorc},
	2={mark=o, \colorc},
	3={mark=diamond, \colorc},
	4={mark=pentagon, \colorc}
	}, 
	scatter src=explicit symbolic, mark size=\msimg,]
	table[select coords between index={10}{14},x index=0,y index=1, meta=class] \toponespace;
	
\addplot[scatter, color=\colorc,
	font=\legsize,
	nodes near coords,
	only marks,
	mark=x,
	black,
	scatter src=explicit symbolic,]
	table[select coords between index={15}{18},x index=0,y index=1, meta=class] \toponespace;
	
\addplot[scatter, color=black,
	font=\legsize,
	nodes near coords,
	only marks,
	mark=+,
	scatter src=explicit symbolic,]
	table[select coords between index={19}{22},x index=0,y index=1] \toponespace;

\addplot[dashed, gray] coordinates {(2.5,69.6) (1e3,69.6)};

\nextgroupplot[xmode=log, xlabel={multiplications}, ylabel={top-5 acc. [\%]},
        legend pos=south east, nodes near coords align=below right, xmin=3.5e6, xmax=3e9,
        y label style={at={(axis description cs:0.2,.5)}}]
\addplot[scatter, color=\colora,
	scatter/classes={ 
	0={mark=triangle, \colora},
	1={mark= square,\colora},
	2={mark=o, \colora},
	3={mark=diamond, \colora},
	4={mark=pentagon, \colora}
	},  
	scatter src=explicit symbolic, mark size=\msimg,]
	table[select coords between index={0}{4},x index=0,y index=1, meta=class] \topfivemul;

\addplot[scatter, color=\colora,
	scatter/classes={ 
	0={mark=square*, \colora},
	1={mark=*,\colora}
	}, 
	scatter src=explicit symbolic, mark size=\msimg, only marks]
	table[select coords between index={0}{0},x index=0,y index=1, meta=class] \topfivemuldist;
	
\addplot[scatter, color=\colorb,
	scatter/classes={ 
	0={mark=triangle,\colorb},
	1={mark=square, \colorb},
	2={mark=o, \colorb},
	3={mark=diamond, \colorb},
	4={mark=pentagon, \colorb}
	}, 
	scatter src=explicit symbolic, mark size=\msimg,]
	table[select coords between index={5}{9},x index=0,y index=1, meta=class] \topfivemul;

\addplot[scatter, color=\colorb,
	scatter/classes={ 
	0={mark=square*, \colorb},
	1={mark=*,\colorb}
	}, 
	scatter src=explicit symbolic, mark size=\msimg, only marks]
	table[select coords between index={1}{2},x index=0,y index=1, meta=class] \topfivemuldist;
	
\addplot[scatter, color=\colorc,
	scatter/classes={ 
	0={mark=triangle, \colorc},
	1={mark=square, \colorc},
	2={mark=o, \colorc},
	3={mark=diamond, \colorc},
	4={mark=pentagon, \colorc}
	}, 
	scatter src=explicit symbolic, mark size=\msimg,]
	table[select coords between index={10}{14},x index=0,y index=1, meta=class] \topfivemul;
	
\addplot[scatter, color=\colorc,
	font=\legsize,
	nodes near coords,
	only marks,
	mark=x,
	black,
	scatter src=explicit symbolic,]
	table[select coords between index={15}{18},x index=0,y index=1, meta=class] \topfivemul;
	
\addplot[scatter, color=black,
	font=\legsize,
	nodes near coords,
	only marks,
	mark=+,
	scatter src=explicit symbolic,]
	table[select coords between index={19}{22},x index=0,y index=1] \topfivemul;

\addplot[dashed, gray] coordinates {(2e6,89.2) (1e10,89.2)};

\nextgroupplot[xmode=log, xlabel={additions}, nodes near coords align=below right, xmin=4.2e8, xmax=1.3e10, yticklabels={,,}]
\addplot[scatter, color=\colora,
	scatter/classes={ 
	0={mark=triangle, \colora},
	1={mark= square,\colora},
	2={mark=o, \colora},
	3={mark=diamond, \colora},
	4={mark=pentagon, \colora}
	},  
	scatter src=explicit symbolic, mark size=\msimg,]
	table[select coords between index={0}{4},x index=0,y index=1, meta=class] \topfiveadd;

\addplot[scatter, color=\colora,
	scatter/classes={ 
	0={mark=square*, \colora},
	1={mark=*,\colora}
	}, 
	scatter src=explicit symbolic, mark size=\msimg, only marks]
	table[select coords between index={0}{0},x index=0,y index=1, meta=class] \topfiveadddist;

\addplot[scatter, color=\colorb,
	scatter/classes={ 
	0={mark=triangle,\colorb},
	1={mark=square, \colorb},
	2={mark=o, \colorb},
	3={mark=diamond, \colorb},
	4={mark=pentagon, \colorb}
	}, 
	scatter src=explicit symbolic, mark size=\msimg,]
	table[select coords between index={5}{9},x index=0,y index=1, meta=class] \topfiveadd;

\addplot[scatter, color=\colorb,
	scatter/classes={ 
	0={mark=square*, \colorb},
	1={mark=*,\colorb}
	}, 
	scatter src=explicit symbolic, mark size=\msimg, only marks]
	table[select coords between index={1}{2},x index=0,y index=1, meta=class] \topfiveadddist;	
	
\addplot[scatter, color=\colorc,
	scatter/classes={ 
	0={mark=triangle, \colorc},
	1={mark=square, \colorc},
	2={mark=o, \colorc},
	3={mark=diamond, \colorc},
	4={mark=pentagon, \colorc}
	}, 
	scatter src=explicit symbolic, mark size=\msimg,]
	table[select coords between index={10}{14},x index=0,y index=1, meta=class] \topfiveadd;
	
\addplot[scatter, color=\colorc,
	font=\legsize,
	nodes near coords,
	only marks,
	mark=x,
	black,
	scatter src=explicit symbolic,]
	table[select coords between index={15}{18},x index=0,y index=1, meta=class] \topfiveadd;
	
\addplot[scatter, color=black,
	font=\legsize,
	nodes near coords,
	only marks,
	mark=+,
	scatter src=explicit symbolic,]
	table[select coords between index={19}{22},x index=0,y index=1] \topfiveadd;

\addplot[dashed, gray] coordinates {(3e8,89.2) (2e10,89.2)};

\nextgroupplot[xmode=log, xlabel={model size [MB]}, nodes near coords align=below left, xmin=7,xmax=400,yticklabels={,,},
legend entries={
	   {\legsize$6$},
            {\legsize$4$},
            {\legsize$2$},
            {\legsize$1$},
            {\legsize$\frac12$}
        }, legend style={at={(-0.6,-0.32)}, anchor=north,legend columns=-1, text=black},legend image post style={black}]
\addplot[scatter, color=\colora,
	scatter/classes={ 
	0={mark=triangle, \colora},
	1={mark= square,\colora},
	2={mark=o, \colora},
	3={mark=diamond, \colora},
	4={mark=pentagon, \colora}
	},  
	scatter src=explicit symbolic, mark size=\msimg,]
	table[select coords between index={0}{4},x index=0,y index=1, meta=class] \topfivespace;
	
\addplot[scatter, color=\colora,
	scatter/classes={ 
	0={mark=square*, \colora},
	1={mark=*,\colora}
	}, 
	scatter src=explicit symbolic, mark size=\msimg, only marks]
	table[select coords between index={0}{0},x index=0,y index=1, meta=class] \topfivespacedist;

\addplot[scatter, color=\colorb,
	scatter/classes={ 
	0={mark=triangle,\colorb},
	1={mark=square, \colorb},
	2={mark=o, \colorb},
	3={mark=diamond, \colorb},
	4={mark=pentagon, \colorb}
	}, 
	scatter src=explicit symbolic, mark size=\msimg,]
	table[select coords between index={5}{9},x index=0,y index=1, meta=class] \topfivespace;

\addplot[scatter, color=\colorb,
	scatter/classes={ 
	0={mark=square*, \colorb},
	1={mark=*,\colorb}
	}, 
	scatter src=explicit symbolic, mark size=\msimg, only marks]
	table[select coords between index={1}{2},x index=0,y index=1, meta=class] \topfivespacedist;

\addplot[scatter, color=\colorc,
	scatter/classes={ 
	0={mark=triangle, \colorc},
	1={mark=square, \colorc},
	2={mark=o, \colorc},
	3={mark=diamond, \colorc},
	4={mark=pentagon, \colorc}
	}, 
	scatter src=explicit symbolic, mark size=\msimg,]
	table[select coords between index={10}{14},x index=0,y index=1, meta=class] \topfivespace;
	
\addplot[scatter, color=\colorc,
	font=\legsize,
	nodes near coords,
	only marks,
	mark=x,
	black,
	scatter src=explicit symbolic,]
	table[select coords between index={15}{18},x index=0,y index=1, meta=class] \topfivespace;
	
\addplot[scatter, color=black,
	font=\legsize,
	nodes near coords,
	only marks,
	mark=+,
	scatter src=explicit symbolic,]
	table[select coords between index={19}{22},x index=0,y index=1] \topfivespace;

\addplot[dashed, gray] coordinates {(2.5,89.2) (1e3,89.2)};

\end{groupplot}
\end{tikzpicture}
\vspace{-0.65cm}
\caption{\label{fig:curves} Top-1 and top-5 validation accuracy of ST-ResNet-18 on ImageNet as a function of the number of multiplications, the number of additions, and model size, along with the values obtained in related works BWN \cite{rastegari2016xnor}, TWN \cite{li2016ternary}, TTQ \cite{zhu2016trained}, ABC-Net-1/2/3/5 \cite{lin2017towards} (``+'' signs, the suffix reflects the ranking according to accuracy), and the full-precision model (FP). The numbers associated with the marker types correspond to the ratio of the number of hidden SP units and output channels, $r/\cout$. Different colors indicate different combinations of output patch size $p$ and number of convolution groups $g$: {\color{\colorb}Blue: $p=2$, $g=1$}; {\color{\colora}green: $p=1$, $g=1$}; {\color{\colorc} red: $p=1$, $g=4$}. Selected models trained with KD are shown with filled markers.}
\vspace{-0.45cm}
\end{figure}

\begingroup
\setlength{\tabcolsep}{0.2mm}
\begin{table}[!th]
\vspace{-0.4cm}
\caption{\label{tab:accimg}Top-1 and top-5 validation accuracy (in \%) of ST-ResNet-18 on ImageNet, for different choices of $r$, $p$, $g$, and with KD.}

  \centering
  \small
  \begin{tabular}{@{\extracolsep{6pt}}lccccccccc@{}}
   \multicolumn{9}{c}{top-1 accuracy} \\
    \toprule
    & \multicolumn{5}{c}{$r$} & \multicolumn{3}{c}{$r$ (KD)}\\ \cline{2-6} \cline{7-9}
$(p, g)$ & $6\cout$ & $4\cout$ & $2\cout$ & $\cout$ & $\frac12 \cout$ & $4\cout$ & $2\cout$ & $\cout$\\[1mm]
\hline
$(1, 1)$ & 67.9 & 67.6 & 67.0 & 64.7 & 62.2 & 68.6 & 67.9 & 66.0\\
$(2, 1)$ & 68.2 & 68.0 & 67.1 & 64.1 & 61.8 & 70.4 & 69.4 & 66.4  \\
$(1, 4)$ & 67.4 & 67.2 & 65.6 & 62.6 & 58.9 & 68.0 & 66.6 & 63.9 \\
    \bottomrule
  \end{tabular}
  
\vspace{1mm}
  \begin{tabular}{@{\extracolsep{6pt}}lccccccccc@{}}
   \multicolumn{9}{c}{top-5 accuracy} \\
    \toprule
    & \multicolumn{5}{c}{$r$} & \multicolumn{3}{c}{$r$ (KD)}\\ \cline{2-6} \cline{7-9}
$(p, g)$ & $6\cout$ & $4\cout$ & $2\cout$ & $\cout$ & $\frac12 \cout$ & $4\cout$ & $2\cout$ & $\cout$\\[1mm]
\hline
$(1, 1)$ & 88.1 & 87.9 & 87.5 & 86.0 & 84.1 & 88.7 & 88.3 & 87.1 \\
$(2, 1)$ & 88.2 & 88.0 & 87.5 & 85.6 & 83.9 & 89.4 & 89.0 & 87.3 \\
$(1, 4)$ & 87.8 & 87.6 & 86.6 & 84.5 & 81.8 & 88.3 & 87.5 & 85.5\\
    \bottomrule
  \end{tabular}
  \vspace{-0.1cm}
\setlength{\tabcolsep}{1.5mm}
\caption{\label{tab:redimg}Reduction in the number of multiplications (in \%) of ST-ResNet-18 compared to the full-precision model, for $224 \times 224$ images.}
  \centering
  \small
  
  \begin{tabular}{lcccccc}
   \multicolumn{6}{c}{red. in multiplications} \\
    \toprule
    & \multicolumn{4}{c}{$r$} \\ \cline{2-6}
$(p, g)$ & $6\cout$ & $4\cout$ & $2\cout$ & $\cout$ & $\frac12 \cout$\\[1mm]
\hline
$(1, 1)$ & 99.01 & 99.29 & 99.56 & 99.73 & 99.79 \\
$(2, 1)$ & 99.63 & 99.70 & 99.77 & 99.83 & 99.85  \\
$(1, 4)$ & 99.01 & 99.29 & 99.56 & 99.73 & 99.79 \\
    \bottomrule
  \end{tabular}

\vspace{-0.5cm}
\end{table}
\endgroup

{\bf Discussion.} All ST-ResNet-18 models that require the same number of multiplications as TWN (those with $r=\cout$, $p=1$, $g=4$; $r=\cout$, $p=1$,  $g=1$; $r=2\cout$, $p=2$, $g=1$) obtain a considerably higher top-1 and top-5 accuracy than TWN. In particular, ST-ResNet-18 with $r=2\cout$, $p=2$, and $g=1$ leads to a 7.0\% improvement in top-1 accuracy. 
Furthermore, ST-ResNet-18 with $r=2\cout$, $p=1$, and $g=1$ outperforms TTQ while using 98.3\% fewer multiplications. ST-ResNet-18 with $r=6\cout$, $p=2$, and $g=1$ incurs a 2.0\% reduction in top-1 accuracy compared to the full-precision model while reducing the number of multiplications by 99.63\%. Our ST-ResNets require fewer multiplications and additions than ABC-Net-1/2/3 while yielding the same accuracy. For $p=2$, our models lead to a reduction in multiplications of at least 50\% compared to the ABC-Net with the same accuracy. Note that TTQ and BWN use considerably more multiplications than ST-ResNet-18, TWN, and ABC-Net as they do not quantize the first convolution layer. 

In contrast to the experiments on CIFAR-10, increasing $p$ from $1$ to $2$ increases the accuracy for fixed $r \geq 2\cout$. A possible explanation for this behavior is that the benefits of the increase in the number of ternary parameters obtained by increasing $p$ outweighs the loss in precision due to the reduction in spatial resolution. This is in accordance with the fact that the images in ImageNet are much larger than those in CIFAR-10, resulting in larger feature maps for most layers.

KD leads to improvements in top-1 accuracy of $1.3$--$3.5$\%, see Table \ref{tab:accimg}. In particular, ST-ResNet-18 with $r=2\cout$, $p=2$, and $g=1$ trained using KD essentially matches the accuracy of the full-precision model. Increasing $r$ to $4\cout$ yields a model that even outperforms the full-precision model. To the best of our knowledge, these models are the first to realize massive reductions in the number of multiplications (over 99.5\%) while maintaining the accuracy of the full-precision model. Note that student models outperforming their teachers were observed before in different contexts \cite{mishra2017apprentice, zhuang2017towards, furlanello2017born}.

For some of the configurations, the reduction in multiplications comes at the cost of a small to moderate increase in the number of additions. We emphasize that this is also the case for Strassen's algorithm (see \eqref{eq:strassen}) and the Winograd filter-based convolution (see \cite{lavin2016fast}, Sec. 4.1). The specific application and target platform will determine what increase in the number of additions is acceptable. 

Finally, in all our image classification experiments the ratio $r/\cout$ is the same for all layers. Since one would expect improvements from allocating more multiplications to layers that require more accurate operations, we also tested a simple way to learn the ratio $r/\cout$ for each layer from data. Specifically, we chose a large $r/\cout$  and applied L1 regularization to the vectors $\tilde a$. However, for a given total multiplication budget this strategy led to lower accuracy in our experiments than just fixing $r/\cout$ for all layers.

\vspace{-0.1cm}
\subsection{Language modeling} \label{sec:languagemodels}

\newcommand{\mslm}{1.0pt}
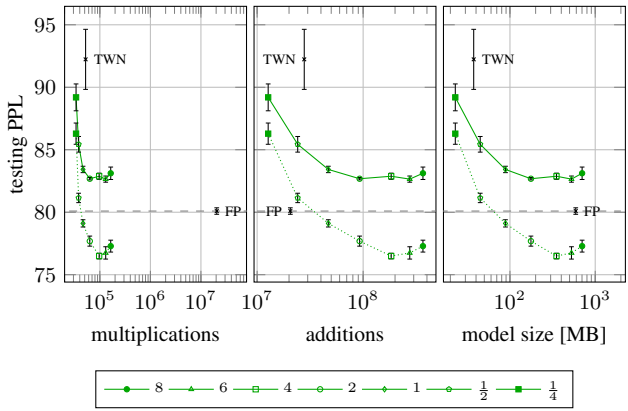
\begin{figure}[t]
\centering
\hspace{-0.28cm}
\begin{tikzpicture}[every node/.style={scale=0.9}] \begin{groupplot}[group style={group size=3 by 2, horizontal sep=0.1cm},height=5.25cm,width=4cm,
enlargelimits=0.1, grid=major, /tikz/font=\small] 
    
\nextgroupplot[xmode=log, xlabel={multiplications}, ylabel={testing PPL},
        nodes near coords align=right, xmin=40000, xmax=4e7, y label style={at={(axis description cs:0.2,.5)}},
        ]
\addplot[
	scatter, color=\colora,
	scatter/classes={
	0={mark=*, \colora},
	1={mark=triangle, \colora},
	2={mark= square,\colora},
	3={mark=o, \colora},
	4={mark=diamond, \colora},
	5={mark=pentagon, \colora},
	6={mark=square*, \colora}
	}, 
	scatter src=explicit symbolic, mark size=\mslm,
	error bars/.cd, y dir=both,y explicit, error bar style={color=black}]
	table[select coords between index={0}{6},x index=0,y index=1, meta=class, y error index=2] \ppwmul;
	
\addplot[
	scatter, color=\colora,
	scatter/classes={
	0={mark=*, \colora},
	1={mark=triangle, \colora},
	2={mark= square,\colora},
	3={mark=o, \colora},
	4={mark=diamond, \colora},
	5={mark=pentagon, \colora},
	6={mark=square*, \colora}
	},
	scatter src=explicit symbolic, mark size=\mslm, densely dotted, mark options={solid},
	error bars/.cd, y dir=both,y explicit, error bar style={color=black, solid},]
	table[select coords between index={0}{6},x index=0,y index=1, meta=class, y error index=2] \ppwmuldist; 
\addplot[scatter, color=\colora,
	font=\legsize,
	nodes near coords,
	only marks,
	mark=x,
	black,
	scatter src=explicit symbolic, mark size=\mslm,
	error bars/.cd, y dir=both,y explicit, error bar style={color=black}]
	table[select coords between index={7}{9},x index=0,y index=1, meta=class, y error index=2] \ppwmul;

\addplot[dashed, gray] coordinates {(12000,80.094) (1e8,80.094)};

\nextgroupplot[xmode=log, xlabel={additions}, nodes near coords align=left, yticklabels={,,}, xmin=1.3e7,xmax=3.5e8]

\addplot[scatter, color=\colora,
	scatter/classes={
	0={mark=*, \colora},
	1={mark=triangle, \colora},
	2={mark= square,\colora},
	3={mark=o, \colora},
	4={mark=diamond, \colora},
	5={mark=pentagon, \colora},
	6={mark=square*, \colora}
	}, 
	scatter src=explicit symbolic, mark size=\mslm,
	error bars/.cd, y dir=both,y explicit, error bar style={color=black}]
	table[select coords between index={0}{6},x index=0,y index=1, meta=class, y error index=2] \ppwadd;
	
\addplot[scatter, color=\colora,
	scatter/classes={
	0={mark=*, \colora},
	1={mark=triangle, \colora},
	2={mark= square,\colora},
	3={mark=o, \colora},
	4={mark=diamond, \colora},
	5={mark=pentagon, \colora},
	6={mark=square*, \colora}
	},
	scatter src=explicit symbolic, mark size=\mslm, densely dotted, mark options={solid},
	error bars/.cd, y dir=both,y explicit, error bar style={color=black, solid}]
	table[select coords between index={0}{6},x index=0,y index=1, meta=class, y error index=2] \ppwadddist;

\addplot[scatter, color=\colora,
	font=\legsize,
	nodes near coords,
	only marks,
	mark=x,
	black,
	scatter src=explicit symbolic, mark size=\mslm,
	error bars/.cd, y dir=both,y explicit, error bar style={color=black}]
	table[select coords between index={7}{9},x index=0,y index=1, meta=class, y error index=2] \ppwadd;
	
\addplot[dashed, gray] coordinates {(1e7,80.094) (5e8,80.094)};

\nextgroupplot[xmode=log, xlabel={model size [MB]}, nodes near coords align=right, xmax=1500, xmin=25,
legend entries={
        {\legsize$8$},
	   {\legsize$6$},
            {\legsize$4$},
            {\legsize$2$},
            {\legsize$1$},
            {\legsize$\frac12$},
            {\legsize$\frac14$}
        }, yticklabels={,,}, 	, legend style={at={(-0.6,-0.32)}, anchor=north,legend columns=-1},]

\addplot[scatter, color=\colora,
	scatter/classes={
	0={mark=*, \colora},
	1={mark=triangle, \colora},
	2={mark= square,\colora},
	3={mark=o, \colora},
	4={mark=diamond, \colora},
	5={mark=pentagon, \colora},
	6={mark=square*, \colora}
	}, 
	scatter src=explicit symbolic, mark size=\mslm,
	error bars/.cd, y dir=both,y explicit, error bar style={color=black}]
	table[select coords between index={0}{6},x index=0,y index=1, meta=class, y error index=2] \ppwspace;
	
\addplot[scatter, color=\colora,
	scatter/classes={
	0={mark=*, \colora},
	1={mark=triangle, \colora},
	2={mark= square,\colora},
	3={mark=o, \colora},
	4={mark=diamond, \colora},
	5={mark=pentagon, \colora},
	6={mark=square*, \colora}
	}, 
	scatter src=explicit symbolic, mark size=\mslm, densely dotted, mark options={solid},
	error bars/.cd, y dir=both,y explicit, error bar style={color=black, solid}]
	table[select coords between index={0}{6},x index=0,y index=1, meta=class, y error index=2] \ppwspacedist;

\addplot[scatter, color=\colora,
	font=\legsize,
	nodes near coords,
	only marks,
	mark=x,
	black,
	scatter src=explicit symbolic, mark size=\mslm,
	error bars/.cd, y dir=both,y explicit, error bar style={color=black}]
	table[select coords between index={7}{9},x index=0,y index=1, meta=class, y error index=2] \ppwspace;

\addplot[dashed, gray] coordinates {(15,80.094) (2e3,80.094)};

\end{groupplot}
\end{tikzpicture}
\vspace{-0.2cm}
\caption{\label{fig:curvesLM} Testing PPL (averaged over $5$ runs) for ST-LM as a function of the number of operations and model size, along with the values obtained for TWN quantization \eqref{eq:quantapprox}, and the full-precision model (FP). Solid line: Without KD; dotted line: With KD. The numbers associated with the marker types correspond to the ratio of the number of hidden SP units and hidden units, $r/\nout$.}
\vspace{-0.5cm}
\end{figure}

We apply our method to the character-level language model described in \cite{kim2016character} and evaluate it on the English Penn Treebank (PTB with word vocabulary size $10$k, character vocabulary size $51$, $1$M training tokens, standard train-validation-test split, see \cite{kim2016character}) \cite{marcus1993building}. We use the large version of the model from \cite{kim2016character} which is composed of a convolution layer with 1100 filters (applied to a character-level representation of the words, without aggregation across channels), followed by a 2-layer highway network with 1100 hidden units, feeding into a 2-layer LSTM network with 650 hidden units (see Table 2 in \cite{kim2016character} for more details). We obtain Strassen language models (ST-LMs) by replacing the convolution layer and all fully connected layers (both within the LSTM and the output/decode layer) with SPNs. $r$ is set to the number of filters for the convolution layer and is parametrized as $r(\kappa)=\kappa \cdot \nout$ for the fully connected layers, where $\nout$ is the number of hidden units. For the output/decode layer we use $r(\kappa)=\kappa \cdot 2000$. 

All models are trained for $40$ epochs using SGD with mini-batch size $20$ and initial learning rate $2$, multiplying the learning rate by $0.5$ when the validation perplexity per word (PPL; c.f. Eq. (9) in \cite{kim2016character}) decreases by less than $0.5$ per epoch (a similar schedule was used in \cite{kim2016character}). 
Although the ST-LMs train stably with quantization from scratch, we train them for 20 epochs with full-precision weights before activating quantization for $W_b$ and $W_c$, which leads to slightly lower validation PPLs. As a baseline, we consider the TWN quantization scheme \eqref{eq:quantapprox} and apply it to all layers of the language model. As we observed a somewhat higher variability in the validation performance than for the image classification experiments, we train each quantized model 5 times and report the average testing PPL.

In Figure \ref{fig:curvesLM}, we plot the average testing PPL of our ST-LMs for different $r$ as a function of the number of operations and model size, with and without KD. Table \ref{tab:redLM} in Appendix \ref{sec:addres} shows the reduction in the number of operations and model size compared to the full-precision model.

{\bf Discussion.} Our ST-LM models reduce the number of multiplications by over 99\% compared to the full-precision model, while incurring an increase in testing PPL of only $3$--$4$\%. The PPL obtained via TWN quantization clearly exceeds that of all considered ST-LMs. The ST-LM model with $r=\nout$ requires roughly the same number of multiplications as the TWN model but has a 7.4\% lower testing PPL. KD leads to a significant reduction in testing PPL. The distilled ST-LMs outperform the teacher model for $r \geq \nout$. To our knowledge, our models are the first to obtain such massive reductions (over 99.5\% for  $r\leq4\nout$) in the number of multiplications while maintaining the PPL of the full-precision model. We observed that KD applied to the teacher model also reduces its testing PPL to values comparable to that of the compressed models with KD for $r \geq \nout$ (see \cite{furlanello2017born} for more exploration of this phenomenon). On the other hand, KD considerably increases the testing PPL for TWN.

There are only few prior works on compressing sequence models in a \emph{multiplication-reducing} fashion \cite{he2016effective, hubara2016quantized}. For single-layer LSTM and GRU language models with binary weights \citet{he2016effective} report an increase in PPL of 70\% and more compared to the full-precision model, whereas \citet{hubara2016quantized} observed divergence for a single-layer LSTM model with binary weights, but report small degradations for 4-bit weight and activation quantization.

\vspace{-0.1cm}
\section{Conclusion and Future Work}

\label{sec:conclusion}
We proposed and evaluated a versatile framework to learn fast approximate matrix multiplications for DNNs end-to-end. We found that our method  leads to the same or higher accuracy compared to existing methods while using significantly fewer multiplications. By leveraging KD we were able to train models that incur no loss in predictive performance despite reducing the number of multiplications by over 99.5\%. 
A natural next step is to incorporate activation quantization into the proposed method. In addition, it will be interesting to see how the theoretical gains reported here translate into actual energy savings and runtime speedups on specialized hardware such as FPGAs and ASICs.

\section*{Acknowledgment} 

The authors would like to thank Eirikur Agustsson, Helmut B\"olcskei, Lukas Cavigelli, Asmus Hetzel, Risi Kondor, Andrew Lavin, Michael Lerjen, Zachary Lipton, Weitang Liu, Andrea Olgiati, John Owens, Sheng Zha, and Zhi Zhang for inspiring discussions and comments. This work was supported by the ``AWS Cloud Credits for Research'' program.

\bibliographystyle{icml2018}
\bibliography{refs.bib}

\clearpage

\appendix

\section{Additional Results on CIFAR-10} \label{sec:vggcifar}

We apply our method to the 7-layer VGG-inspired architecture previously considered in \cite{courbariaux2015binaryconnect, li2016ternary} (see (\citet{li2016ternary}, Sec. 3) for a detailed description of the architecture) and evaluate it on CIFAR-10. We vary $r$ and $p$ for convolution layers and fix $r=1024$ for the fully connected layer with $1024$ units feeding into the softmax layer. The same hyper parameters and schedules as for ST-ResNet-20 are used for training, see Sec. \ref{sec:cifar}. Table \ref{tab:vggcifar} shows the testing accuracy for different $r$ and $p$. Our method achieves the same or higher accuracy than TWN \cite{li2016ternary} for $p=1$. The impact of increasing $p$ on testing accuracy is analogous to that observed for ST-ResNet-20. Reducing $r$ seems to reduce testing accuracy to a smaller extent than for ST-ResNet-20. A possible reason for this could be that the considered VGG architecture has considerably wider layers ($128$ to $512$ channels) than ResNet-20 ($16$ to $64$ channels).

\begin{table}[h]
\vspace{-0.4cm}
  \caption{\label{tab:vggcifar}Testing accuracy (in \%) of the 7-layer VVG model from \cite{courbariaux2015binaryconnect, li2016ternary} compressed by our method, on CIFAR-10.}
%   \vspace{2mm}
  \centering
  \small
  \begin{tabular}{lccccc}
  \multicolumn{5}{c}{testing accuracy} \\
    \toprule
    & \multicolumn{4}{c}{$r$} \\ \cline{2-5}
$p$ & $\cout$ & $\frac{3}{4}\cout$ & $\frac{1}{2}\cout$ & $\frac{1}{4}\cout$ \\[1mm]
\hline
$1$ & 93.17 & 93.19 & 92.39 & 92.50 \\
$2$ & 91.87 & 91.44 & 91.39 & 89.66  \\
$4$ & 88.08 & 88.40 & 87.65 & 87.05 \\
    \bottomrule
  \end{tabular}
  \vspace{-0.4cm}
\end{table}

\section{Additional Results for Language Modeling}

To assess the generalization of the ST-LM models described in Section~\ref{sec:languagemodels}, we apply the FP and ST-LM (with $r=2\nout$) models to Wikitext-2\footnote{Available at {\url{https://www.salesforce.com/products/einstein/ai-research/the-wikitext-dependency-language-modeling-dataset/}}.} (word vocabulary size $33$k and $2$M training tokens) without changing hyper parameters and obtain a testing PPL of 90.07, 97.40, and 87.72 for the FP model, the ST-LM model, and the ST-LM model with KD, respectively. The PPL of the FP model ($19$M parameters) is comparable to that of the variational dropout LSTM (VD-LSTM-RE, $22$M parameters) from \cite{inan2016tying}. While the gap between the FP and ST-LM model is larger than for PTB, the ST-LM model with distillation outperforms the FP model similarly as for PTB.

\vspace{-0.2cm}
\section{Application to 2D convolution: Pseudocode} \label{sec:2dconvpseudo}

In this section, we provide provide pseudocode to facilitate the implementation of the proposed framework for 2D convolutions with $k\times k$ kernels (see Section~\ref{sec:2dconv}) in popular deep learning software packages. Let \texttt{W\_B}, \texttt{a\_tilde}, and \texttt{W\_C} be variables associated with tensors of dimensions $r \times \cin \times (p-1+k) \times (p-1+k)$, $1 \times r \times 1 \times 1$, and $r \times \cout \times p \times p$, respectively. Denote the standard 2D convolution and transposed 2D convolution operations with input \texttt{data}, filter tensor \texttt{weights}, \texttt{in\_channels} input channels, \texttt{out\_channels} output channels, kernel size \texttt{kernel\_size}, and stride \texttt{stride} by \texttt{Conv2d} and \texttt{ConvTranspose2d}. Let \texttt{Multiply} be the broadcasted element-wise multiplication of \texttt{weights} with \texttt{data} and designate the function implementing the quantization scheme described in \eqref{eq:quantapprox} by \texttt{Quantize}. Then, the forward pass (during training) through a compressed 2D convolution for an input tensor \texttt{in\_data} of dimensions $b \times \cin \times W \times H$ is given by
\vspace{-0.1cm}
\texttt{
\begin{tabbing}
W\_B = Quantize(W\_B) \\
W\_C = Quantize(W\_C) \\
conv\_out = \= Conv2d( \\
\>~~~~data=in\_data, \\
\>~~~~weights=W\_B, \\
\>~~~~in\_channels=$\cin$, \\
\>~~~~out\_channels=$r$, \\
\>~~~~kernel\_size=$p-1+k$, \\
\>~~~~stride=$p$, \\
\>~~~~groups=$g$) \\
mul\_out = Multiply(\\
\>~~~~data=conv\_out, \\
\>~~~~weights=a\_tilde) \\
out\_data = ConvTranspose2d(\\
\>~~~~data=mul\_out, \\
\>~~~~weights=W\_C, \\
\>~~~~in\_channels=$r$, \\
\>~~~~out\_channels=$\cout$, \\
\>~~~~kernel\_size=$p$, \\
\>~~~~stride=$p$)
\end{tabbing}
\vspace{-0.1cm}
}
At inference time, \texttt{Conv2d} and \texttt{ConvTranspose2d} can be replaced with specialized convolution operations exploiting the fact that \texttt{W\_B} and \texttt{W\_C} are ternary. To compute the backward pass, the backpropagation algorithm \cite{rumelhart1986learning} is applied to the sequence of operations in the forward pass, ignoring the \texttt{Quantize} operations. We found it beneficial to perform batch normalization \cite{ioffe2015batch} after the \texttt{Conv2d} operation.

\vspace{-0.2cm}
\section{Additional Tables} \label{sec:addres}

\setlength{\extrarowheight}{1mm}
\begingroup
\begin{table}[h!]
\vspace{-0.4cm}
 \caption{\label{tab:compcifarapp}Reduction (in \%) in the number of additions obtained by our method for ResNet-20 on CIFAR-10.}
 \vspace{2mm}
\centering
\small
\begin{tabular}{lcccc}
      \multicolumn{5}{c}{red. in additions} \\
    \toprule
    & \multicolumn{4}{c}{$r$} \\ \cline{2-5}
 $p$ & $\cout$ & $\frac{3}{4}\cout$ & $\frac{1}{2}\cout$ & $\frac{1}{4}\cout$ \\[1mm]
\hline
$1$ & -13.904 & 14.435 & 42.774 & 71.112 \\
$2$ & 39.123 & 54.205 & 69.287 & 84.369 \\
$4$ & 57.550 & 68.025 & 78.501 & 88.976 \\
  \bottomrule
  \end{tabular}
\end{table}
\vspace{-0.2cm}
\endgroup

\begingroup
\setlength{\tabcolsep}{1.2mm}
\begin{table*}[t]
\vspace{-0.3cm}
\caption{\label{tab:redLM} Testing PPL and reduction (in \%) in the number of operations as well as model size for ST-LM compared to the full-precision model. We also report the reductions for the model compressed with TWN quantization \eqref{eq:quantapprox}.}
  \centering
  \small
  \begin{tabular}{lrrrrrrrrr}
\\    \toprule
    & \multicolumn{7}{c}{ST-LM, r} \\ \cline{2-8}
& $8\nout$ & $6\nout$ & $4\nout$ & $2\nout$ & $\nout$ & $\frac12 \nout$ & $\frac14 \nout$ & TWN\\[1mm]
\hline
testing PPL & 83.118 & 82.65 & 82.88 & 82.68 & 83.42 & 85.44 & 89.19 & 92.23 \\
testing PPL (dist.) & 77.29 & 76.74 & 76.49 &  77.69 & 79.11 & 81.14 & 86.29 & - \\
\hline
multiplications & 
99.20 &
99.37 &
99.53 &
99.69 &
99.77 &
99.82 &
99.84 &
99.75 \\
additions & 
-1682.23 &
-1238.24 &
-794.28 &
-350.31 &
-128.33 &
-17.34 &
38.21 &
-35.27 \\
model size &
-18.76 &
10.89 &
40.54 &
70.19 &
85.01 &
92.42 &
96.13 &
93.65\\
    \bottomrule
  \end{tabular}
  \end{table*}
  \endgroup

\begin{table}[t]
 \caption{\label{tab:redimgapp}Reduction in the number of additions and model size (in~\%) of ST-ResNet-18 compared to the full-precision model, for $224\times224$ images.}
  \centering
  \small
\vspace{2mm}
  \begin{tabular}{lcccccc}
   \multicolumn{6}{c}{red. in additions} \\
    \toprule
    & \multicolumn{4}{c}{$r$} \\ \cline{2-6}
$(p, g)$ & $6\cout$ & $4\cout$ & $2\cout$ & $\cout$ & $\frac12 \cout$\\[1mm]
\hline
$(1, 1)$ & -596.76 & -364.56 & -132.36 & -16.32 & 41.73 \\
$(2, 1)$ & -288.57 & -159.10 & -29.63 & 35.05 & 67.41  \\
$(1, 4)$ & -181.51 & -87.73 & 6.05 & 52.89 & 76.33 \\
    \bottomrule
  \end{tabular}
   \vspace{2mm}
   
  \begin{tabular}{lcccccc}
   \multicolumn{6}{c}{red. in model size} \\
    \toprule
    & \multicolumn{4}{c}{$r$} \\ \cline{2-6}
$(p, g)$ & $6\cout$ & $4\cout$ & $2\cout$ & $\cout$ & $\frac12 \cout$\\[1mm]
\hline
$(1, 1)$ & 53.87 & 67.76 & 81.64 & 92.16 & 95.63 \\
$(2, 1)$ & 3.07 & 33.89 & 64.71 & 83.69 & 91.40  \\
$(1, 4)$ & 80.31 & 85.38 & 90.45 & 96.56 & 97.83 \\
    \bottomrule
  \end{tabular}
\end{table}  

\begin{table}[h]
\caption{\label{tab:accrelimg} Top-1 and top-5 validation accuracy (in \%) along with the reduction (in \%) in the number of multiplications and model size for BWN \cite{rastegari2016xnor}, TWN \cite{li2016ternary}, TTQ \cite{zhu2016trained}, ABC-Net-3 \cite{lin2017towards}, and ST-ResNet-18-2-2-1 ($r=2\cout$, $p=2$, $g=1$), compared to the full-precision model (for the full-precision model, the absolute quantities are given in parentheses).}
  \centering \small
  \setlength{\tabcolsep}{1.2mm}
  \begin{tabular}{lcccccc}
\\    \toprule
 & top-1 & top-5 & mul. & model size\\[1mm]
\hline
BWN & 60.8 & 83.0 & 93.25 & 92.33 \\
TWN & 61.8 & 84.2 & 99.73 & 93.39  \\
TTQ & 66.6 & 87.2 & 86.66 & 89.20 \\
ABC-Net-3 & 66.2 & 86.7 & 99.42 & 49.39 \\
ST-ResNet-18-2-2-1 & 67.1 & 87.5 & 99.77 & 125.89 \\
\hline
full-precision & 69.6 & 89.2& ($1.82\cdot10^9$) & (356.74 MB) \\     \bottomrule
  \end{tabular}
\end{table}

\FloatBarrier

\end{document}